%% file: 00_selfsup_synth_blur_imavis.tex
\documentclass[5p, twocolumn]{elsarticle}  

\usepackage{lineno,hyperref}
\usepackage{amsmath}  
\usepackage{amssymb}  
\input{aitors_usepackages}  

\journal{Image and Vision Computing}

\begin{document}
\input{0_frontmatter_imavis}

\renewcommand*{\today}{August 2, 2019}
\input{1_introduction}
\input{2_blur_generation}
\input{3_model}

\input{4_results}
\input{5_conclusions}
\input{6_acknowledgements}

\section*{References}
\bibliography{longstrings,selfsup_synth_blur_refs}

\end{document}

%% file: aitors_usepackages.tex
\usepackage{booktabs} 

\usepackage{etoolbox,siunitx}
\robustify\bfseries

\usepackage{tablefootnote} 

\usepackage[dvipsnames]{xcolor} 

\newcommand{\colfirst}{\textcolor{OliveGreen}}
\newcommand{\colsecond}{\textcolor{RoyalBlue}}
\newcommand{\colthird}{\textcolor{YellowOrange}}

\usepackage[caption=false,font=footnotesize]{subfig}

\usepackage{blindtext}  

\usepackage[inline]{enumitem} 

\makeatletter
\DeclareRobustCommand\onedot{\futurelet\@let@token\@onedot}
\def\@onedot{\ifx\@let@token.\else.\null\fi\xspace}

\def\eg{\emph{e.g}\onedot} 
\def\ie{\emph{i.e}\onedot}

\def\etal{\emph{et al}\onedot}
\makeatother

%% file: 0_frontmatter_imavis.tex
\begin{frontmatter}

	\title{Self-supervised Blur Detection from Synthetically Blurred Scenes}
	
	\author[tecnaliaaddress,cvcaddress]{Aitor~Alvarez-Gila\corref{mycorrespondingauthor}}
	\cortext[mycorrespondingauthor]{Corresponding author}
	\ead{aitor.alvarez@tecnalia.com}
	
	\author[etsmaddress]{Adrian~Galdran}
	\ead{adrian.galdran-cabello.1@ens.etsmtl.ca}	
	
	\author[tecnaliaaddress]{Estibaliz~Garrote}
	\ead{estibaliz.garrote@tecnalia.com}	

	\author[cvcaddress]{Joost~van~de~Weijer}
	\ead{joost@cvc.uab.es}
	
	\address[tecnaliaaddress]{TECNALIA, Derio, Spain}
	\address[cvcaddress]{CVC - Universitat Aut\`onoma de Barcelona, Barcelona, Spain}
	\address[etsmaddress]{\'Ecole de Technologie Sup\'erieure, Montr\'eal, Canada}
	
	\begin{abstract}
		Blur detection aims at segmenting the blurred areas of a given image.
		Recent deep learning-based methods approach this problem by learning an end-to-end mapping between the blurred input and a binary mask representing the localization of its blurred areas. 
		Nevertheless, the effectiveness of such deep models is limited due to the scarcity of datasets annotated in terms of blur segmentation, as blur annotation is labour intensive.
		In this work, we bypass the need for such annotated datasets for end-to-end learning, and instead rely on object proposals and a model for blur generation in order to produce a dataset of synthetically blurred images.
		This allows us to perform self-supervised learning over the generated image and ground truth blur mask pairs using CNNs, defining a framework that can be employed in purely self-supervised, weakly supervised or semi-supervised configurations.
		Interestingly, experimental results of such setups over the largest blur segmentation datasets available show that this approach achieves state of the art results in blur segmentation, even without ever observing any real blurred image.
	\end{abstract}
	
	\begin{keyword}
		blur\sep defocus\sep motion\sep deep learning\sep self-supervised learning\sep synthetic
		\MSC[2010] 68T45\sep  62H35 
	\end{keyword}

\end{frontmatter}

%% file: 1_introduction.tex

\section{Introduction}\label{sec:intro}

\input{fig_abstract}
Image blur is a phenomenon that degrades the definition of an image, producing a loss of detail in the affected regions. 
The two main causes leading to a fully or partially blurred image are 
\begin{enumerate*}[label=(\roman*)]
	\item defocus blur, which is inherent to a wide aperture optical image capturing device that projects scene points that are away from the focus plane onto a non-punctual circle of confusion on the sensor and 
	\item motion blur, which is caused by the movement of either the camera (\ie camera shake) or the imaged objects during camera exposure time. 
\end{enumerate*}
Even if both defocus and object-motion blurs are sometimes sought-after as part of a creative photographic process (\eg to pop-out the subject or to evoking a sense of motion, respectively), most of the time blur is considered as an undesired effect~\cite{zhang_learning_2018} or image artifact.

In any case, trying to localize blurred regions within an image (or, equivalently, achieving a segmentation in terms of blurry/unblurred parts)  is a useful task with a wide range of applications in computational photography, e.g. defocus blur magnification~\cite{bae_defocus_2007, golestaneh_spatially-varying_2017}, image deblurring~\cite{zhang_spatially_2016,pan_soft-segmentation_2016,shi_just_2015,gast_parametric_2016}, or camera focus point or depth of focus estimation~\cite{golestaneh_spatially-varying_2017}.
In addition, due the underlying correlation between blurred  and non-blurred areas within the same image, blur detection has been also applied in general computer vision tasks, \eg depth estimation~\cite{gur_single_2019}, saliency prediction~\cite{ding_improving_2019} or semantic object segmentation~\cite{pertuz_focusaided_2015}. 
However, blurred region segmentation is a challenging task, due to the fundamental ambiguities existing between the out-of-focus pixels and the originally flat regions or smooth edges.
Scale-ambiguity (\ie the difficulty of inferring the level of blur over one single scale~\cite{shi_discriminative_2014}) and the dependence of the perception of sharpness on the image size are additional challenges that affect the performance of current approaches.

A number of previous works have investigated blur localization directly or implicitly from the feature engineering and physical modeling approaches, either taking one single~\cite{tang_defocus_2013,shi_discriminative_2014, golestaneh_spatially-varying_2017} or  multiple~\cite{favaro_shape_2008, zhou_coded_2009, zhou_depth_2010} images as input, and aiming at detecting only one~\cite{zhu_estimating_2013,tang_defocus_2013,shi_just_2015,chen_fast_2016, yi_lbp-based_2016, pang_motion_2016,zhu_efficient_2016} or both kinds of blur~\cite{liu_image_2008, chakrabarti_analyzing_2010, su_blurred_2011}.
Most of these try to leverage information extracted directly from the intensities~\cite{liu_image_2008}, from the gradients~\cite{zhuo_defocus_2011,su_blurred_2011,pang_classifying_2016,zhu_efficient_2016}, or from transformed domains~\cite{chakrabarti_analyzing_2010,zhu_estimating_2013,tang_defocus_2013,shi_discriminative_2014,tang_spectral_2016}.

More recently, supervised learning-based approaches~\cite{pang_classifying_2016}, and particularly those based on the use of Convolutional Neural Networks (CNN), have shown enormous potential for tackling tasks that require a dense, per-pixel prediction, such as semantic segmentation~\cite{shelhamer_fully_2016,chen_rethinking_2017}, instance segmentation~\cite{he_mask_2017} or crowd counting via density map estimation~\cite{liu_leveraging_2018}. 
Blur segmentation can also be viewed as one of such dense prediction tasks, and several works have already explored this approach, either for predicting both types~\cite{ma_deep_2018a, zhang_learning_2018,kim_defocus_2018} or defocus only blur~\cite{park_unified_2017,zeng_local_2019,zhao_defocus_2018,zhao_defocus_2019}.

Nevertheless, the performance gain obtained by these fully-supervised, CNN-based approaches trained end-to-end is relatively modest when compared to gains in other fields.
The main bottleneck which hinders the full power of CNNs in their application to the field of blur localization is the absence of large enough datasets with pixel-wise annotations.
Several recent efforts have partially addressed this void~\cite{shi_discriminative_2014, zhang_learning_2018, zhao_defocus_2018, zhao_defocus_2019}, but these manual annotations are scarce, labour intensive and costly to acquire.

In the absence of large public sets of annotated data, the generation of synthetic image degradations has been successfully applied for training deep CNNs in recent years.
This approach has been adopted for computer vision tasks as diverse as image super-resolution~\cite{ledig_photo-realistic_2016}, inpainting~\cite{pathak_context_2016}, image quality assessment~\cite{liu_rankiqa:_2017} or perceptual similarity estimation~\cite{zhang_unreasonable_2018}.
It is also a core component of many self-supervised learning methods, which, facing the lack of large sets of annotated data for a certain final task, define a pretext task (\eg inpainting~\cite{pathak_context_2016}, warped image matching~\cite{novotny_self-supervised_2018}, artifact spotting~\cite{jenni_self-supervised_2018}) for which ground truth labels can be automatically derived without manual intervention and be used for feature learning.
A particular case of such scenario would be that of pretext and final tasks being equal.

Synthetic blur generation has also been explored as part of deblurring workflows~\cite{kupyn_deblurgan:_2018,pan_soft-segmentation_2016}, as the blur generation process has been extensively studied, especially within deconvolution approaches for image deblurring.
In the context of blur localization, however, synthetic blur operations have only been applied either over very simple partial blur masks (\eg consisting of a hard rectilinear partition of the image in halves~\cite{zhao_defocus_2019}) or globally over full patches of reduced size extracted from the whole image~\cite{purohit_learning_2018}. 

Our main contribution in this work is the introduction of a deep self-supervised partial blur detection framework, which successfully localizes both defocus and object-motion blur types for a single image without making use of any blur segmentation annotation for training the model. 
Instead, we circumvent the lack of large annotated blur detection image sets by selectively applying procedural synthetic blurring operations to varying regions within images taken from an unrelated non-annotated dataset of natural images.
By controlling the definition of the regions being synthetically blurred, we can automatically generate the associated ground-truth blur masks on the fly.
Fig.~\ref{fig_abstract} illustrates the process.

The framework comprises three different instantiations, defined by i) how those regions are extracted and ii) whether the training is purely based on synthetically blurred images or not:
\begin{enumerate}
	\item A \textbf{self-supervised} approach, in which the regions to be blurred are defined by an object proposal~\cite{krahenbuhl_learning_2015} model, which can generate class-agnostic plausible object masks and thus generalize well across a variety of datasets. 
	This method can extract multiple different blur masks per image, so that the model learns from different image and blur-mask pairs. 
	This allows us to obtain a highly variable training image stream to feed our CNN-based semantic segmentation model at train time. 
	\item A \textbf{weakly-supervised} approach, in which such regions are instead selected from the ground truth labels of any given semantic segmentation image set.
	\item A \textbf{semi-supervised} approach, in which the synthetically generated blurred image and labels from the object proposal method are used in conjunction with real partially blurred images and their respective manual blur-mask annotations in order to augment the given fully supervised blur segmentation dataset. 
\end{enumerate}

By extensively evaluating the learned model on the two largest publicly available pixel-wise annotated real blur localization datasets~\cite{shi_discriminative_2014, zhao_defocus_2018}, we show that our approach generalizes adequately for both blur types, improving the performance of all the considered classic approaches and that of recent fully supervised deep CNN-based ones that require large human-labeled training sets. 
Experimental results show that the proposed solution can be successfully employed to train a deep blur segmentation model, either without the need for any specific blur localization dataset or by making use of a very reduced set of images exhibiting real blur degradations.  
This is especially important for domains other than that of natural RGB images, for which no labeled blur masks exist, such as infrared or multispectral~\cite{chen_multispectral_2015} imagery, medical imaging~\cite{lopez_automated_2013} or text documents~\cite{maheshwari_document_2016}.

%% file: fig_abstract.tex
\begin{figure*}[t!]
\centering
\includegraphics[height=10.9cm, width=\textwidth]{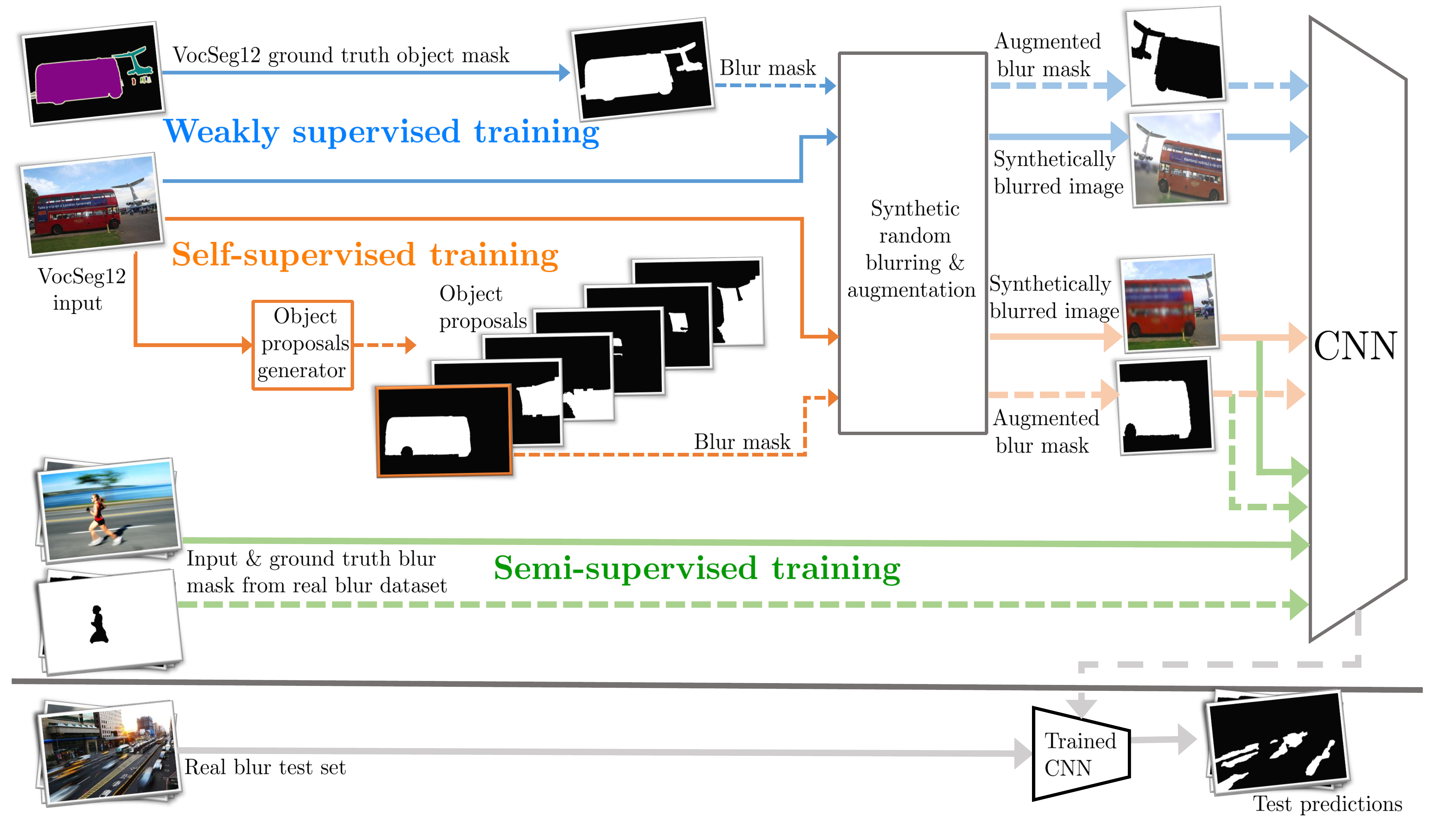}
\caption{General overview of our framework train and testing processes, with each path color representing one of its three possible instantiations \ie self-supervised, weakly-supervised and semi-supervised approaches.}
\label{fig_abstract}
\end{figure*}

%% file: 2_blur_generation.tex
\section{Synthesizing realistic blur}\label{sec:synth}

In this paper, we are interested in learning to localize blur on partially blurred images without using any labeled example for this specific task (in the self-supervised and weakly supervised approaches). 
We therefore switch to the problem of generating plausible blurred scenes from non-blurred images.

\subsection{Blur mask extraction}\label{ssec:blur_mask}

The first step of the process is the determination of the parts of the image that will be subject to the synthetic blurring operation, to which we refer as the \emph{blur mask}.
\input{fig_mcg}

\subsubsection{Semantic object masks as blur masks}\label{sssec:semantic_object_masks}

One possible approach to achieve this is that of using a dataset of images on which several objects have been manually segmented, without any explicit relation to its blur content.
This effectively implies making use of a distant supervision (one grounded on easier to obtain semantic object annotations) for the final blur segmentation task.
We refer to this as our weakly supervised approach.

This kind of labeled data is readily available from datasets generated for semantic segmentation tasks, such as the Pascal VOC Segmentation challenge $2012$ (SegVOC12) \cite{everingham_pascal_2015}.
This dataset includes ground truth annotations of the most prominent objects present in the image (see Fig.~\ref{fig_mcg_gt_obj}) corresponding to $20$ different classes.
Under this setting, at train time, we build the blur mask for each input image by computing the connected components of the largest object present in the corresponding ground truth (Fig.~\ref{fig_mcg_gt0}).

\input{fig_ex}

\subsubsection{Object proposals as blur masks}\label{sssec:object_proposals}

Our purely self-supervised method takes one step ahead by removing the need for manual object segmentation, and replacing it with the inclusion of a class-agnostic object mask proposal generation step.
The goal of object proposal generation methods is, given an input image, to yield a set of either bounding boxes or  segmentation masks that correspond to different object location hypotheses.
Its primary application is serving as a first candidate location filtering stage of two-phase object detection methods, so that more resources can be allocated for the representation and analysis of the resulting subset of regions.

The Multiscale Combinatorial Grouping (MCG) object proposal generation algorithm~\cite{pont-tuset_multiscale_2017} represents one of the most accurate approaches of its kind.
Moreover, although some learning-based steps are involved in its model creation process, its ability to generalize across different datasets renders this method virtually parameter-free.
It yields a ranked set of segmented object proposals after a process that comprises: 
\begin{enumerate*}[label=(\roman*)]
	\item a multi-scale input image segmentation step (based on low level features), 
	\item a rescaling and alignment of the segmentation results, 
	\item the combination of such results onto a merged multi-scale hierarchy of binary spatial image partitions (\ie regions) and 
	\item a final combinatorial grouping stage, which explores the region tree looking for sets of regions that, merged together, are likely to represent complete objects. 
\end{enumerate*}
The resulting set of -hundreds of- proposals is then ranked in accordance to a score representing such likelihood.
During training, we apply MCG to every input image and, at each epoch, we randomly sample a blur mask from the probability distribution given by applying a softmax operation over the mentioned scores.

Fig. \ref{fig_mcg} shows an example of blur mask extraction for both ground truth semantic object mask (b-c) and MCG-based object proposal (e-h) methods.
It is important to note that, should we directly use the blur mask extracted as described in either case, we would be introducing a strong bias in the blur detection training, favoring the prediction of blurred elements in the foreground or background of the images.
This is due to a comparatively significant amount of the extracted blur masks corresponding to foreground objects.
In order to mitigate this and promote the invariance of the model to this respect, we invert the blur mask with a probability $p_{inv}$.

\input{fig_halo}
\subsection{Synthetic blur}\label{ssec:synthetic_blur}
Once the input blur mask has been created, we can now randomly apply different kinds of synthetic blur operations over it.
Given an image $I:\Omega\subset\mathbb{R}^2 \rightarrow \mathbb{R}$ for which the image domain $\Omega$ has been already partitioned into background $\Omega^B$ and foreground $\Omega^F$, represented as the blur mask, we generate a partially blurred version of $I$ by first defining a blur kernel $K$.
Then, the resulting artificially blurred image $I_b$ can be easily obtained by computing:
\begin{equation}
I_b(x,y) = \left\{\begin{array}{lr}
        K*I(x,y), & \text{for } (x,y) \in \Omega^B,\\
        I(x,y), & \text{for } (x,y) \in \Omega^F.
        \end{array}\right.
\end{equation}

The adequate definition of the kernel $K$ is critical in order to accomplish the goal of generating realistic blur. 
While blur coming from situations on which an object is in-focus and the background is defocused can be easily simulated by Gaussian kernels $K_\sigma$ of varying standard deviations $\sigma$, blurs of different natures, \eg motion blur, are harder to emulate. 
Since our aim is to produce fast blurred versions of a given background on the fly, we design a simple pipeline for generating non-linear motion blur:
\begin{enumerate}
\item Build a horizontal line of length $1\times m$ in a discrete domain $\Omega_k \subset \Omega$, obtaining a kernel $K_m$ that defines a linear horizontal motion blur.
\item Rotate it by $\alpha$ degrees, obtaining  a kernel $K_{(m,\alpha)}$ that defines a linear diagonal motion blur.
\item Apply an elastic deformation $\mathcal{E}:\Omega_k\rightarrow\Omega_k$ of the underlying image grid $\Omega_k$ that turns $K_{(m,\alpha)}$ into a non-linear kernel $K_{(m,\alpha,\mathcal{E})}$.
\end{enumerate}
In training time, a random decision of whether to apply defocus blur or motion blur is made for each training image. The definitions of $\sigma$ or $(m,\alpha,\mathcal{E})$  are also randomized by drawing parameters from a suitable range of values.
In this way, the same image can be transformed in infinitely many ways. 

In Fig.~(\ref{fig_ex_a}) we show one image from our Pascal VOC training set. In Figs.~(\ref{fig_ex_c}) and (\ref{fig_ex_d}) the background, as given by Fig.~(\ref{fig_ex_b}), is blurred with two Gaussians of varying standard deviations, whereas in Figs. (\ref{fig_ex_e}--\ref{fig_ex_h}) different non-linear motion blurs are applied. 
It should be noted that there exist more sophisticated mechanisms to simulate non-linear image blurring based on physical considerations~\cite{kupyn_deblurgan:_2018}.
However, for the purposes of this work we prioritize a simple and efficient strategy like the one described above.

Note that, although ideally we would want to use perfectly sharp images to perform the selective blurring, the employed SegVOC12 dataset does indeed contain some pictures exhibiting preexisting partial blur.
This means that the artificially generated ground truth will contain a certain amount of noisy pixel-level labels.
In spite of this, as shown in section~\ref{sec:results}, our framework is able to successfully learn to segment real blur.

\subsection{Removing halo artifacts by inpainting}\label{ssec:inpainting}

The naive approach of blurring the image, and then placing back the unblurred foreground on its original location has some disadvantages.
Namely, this procedure leads to the appearance of halo artifacts around the borders of the foreground, as shown in Figs. (\ref{fig_blur3}) and (\ref{fig_blur4}).
The reason of this problem is that sharp intensity jumps at borders of foreground objects artificially distort image statistics when averaging with a blur kernel around those pixels.

If a model is trained with images containing these halos, it will likely learn to localize blur by simply finding the position of such artifacts.
In order to avoid this situation, we propose a different approach to obtain a blurred-background image.
After extracting the sharp foreground from a given image, we proceed to inpaint the foreground pixels applying the method from~\cite{telea_image_2004}, as shown in Fig. (\ref{fig_blur2}), before blurring the background.
This process allows to remove halo artifacts from the artificially blurred scenes before supplying them to our model, see Figs. (\ref{fig_blur5}) and (\ref{fig_blur6}).

%% file: fig_mcg.tex
\begin{figure}[!htpb]
	\centering
	\subfloat[Input]{\includegraphics[width=0.3\columnwidth]{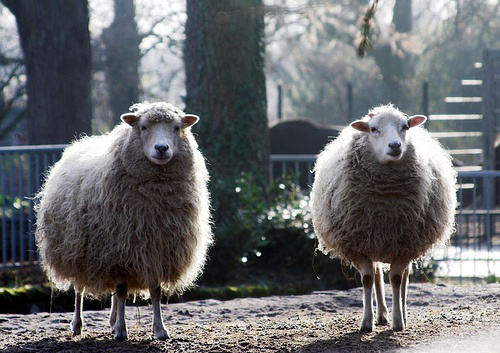}
		\label{fig_mcg_input}}
	\hfil
	\subfloat[VOC objs.]{\includegraphics[width=0.3\columnwidth]{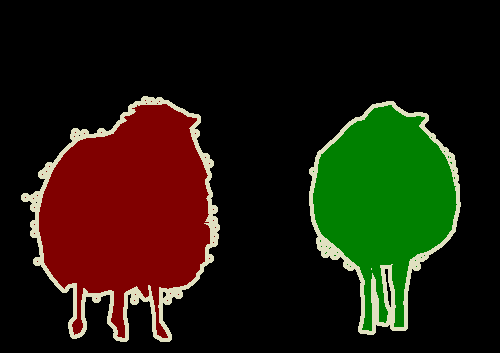}
		\label{fig_mcg_gt_obj}}
	\subfloat[VOC mask]{\includegraphics[width=0.3\columnwidth]{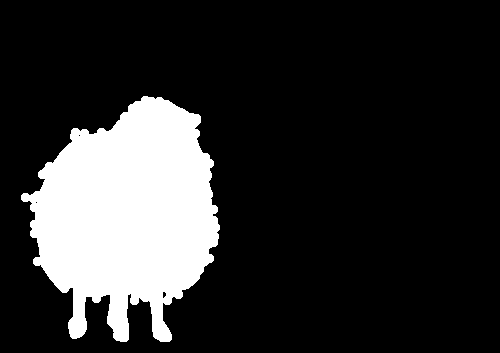}
		\label{fig_mcg_gt0}}
	\hfil
	\subfloat[Scores]{\includegraphics[width=0.3\columnwidth]{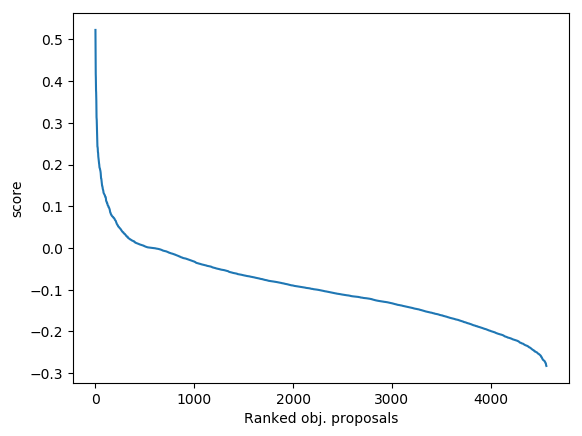}
	\label{fig_mcg_scores}}
	\subfloat[Obj. prop. 1]{\includegraphics[width=0.3\columnwidth]{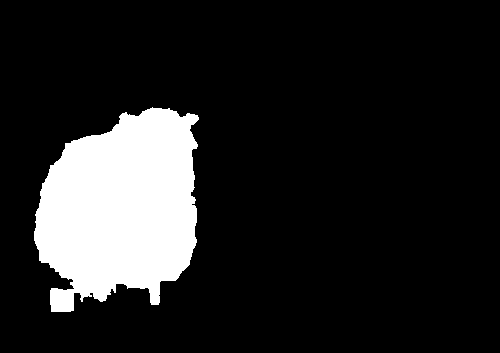}
		\label{fig_mcg_mask0}}
	\hfil
	\subfloat[Obj. prop. 2]{\includegraphics[width=0.3\columnwidth]{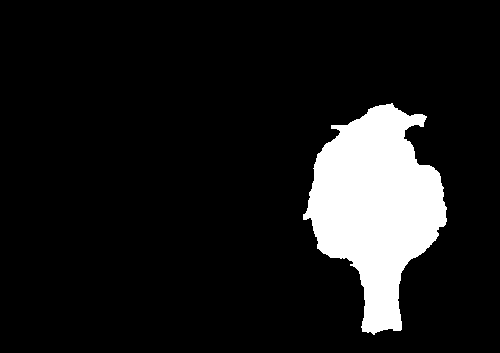}
		\label{fig_mcg_mask1}}
	\hfil
	\subfloat[Obj. prop. 3]{\includegraphics[width=0.3\columnwidth]{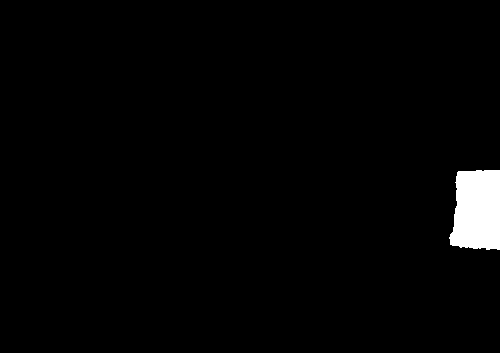}
		\label{fig_mcg_mask2}}
\hfil
	\subfloat[Obj. prop. 4]{\includegraphics[width=0.3\columnwidth]{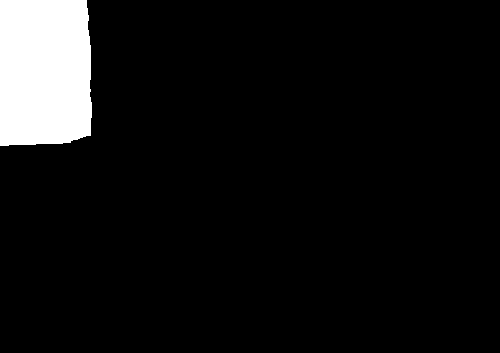}
		\label{fig_mcg_mask3}}
\hfil
	\subfloat[Obj. prop. 5]{\includegraphics[width=0.3\columnwidth]{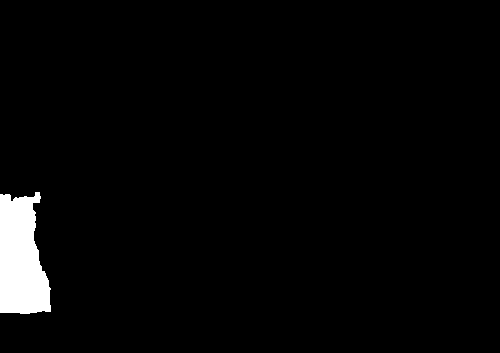}
		\label{fig_mcg_mask4}}
	\caption{Blur mask extraction from an input image (a) of the Pascal VOC 2012 dataset~\cite{everingham_pascal_2015}. (b) Ground truth object masks from the segmentation challenge. (c) Blur mask given by the largest connected component. (d) Sorted scores of objectness given by MCG for this input. (e-i) First five object proposals generated by MCG~\cite{pont-tuset_multiscale_2017}.}
	\label{fig_mcg}
\end{figure}

%% file: fig_ex.tex
\begin{figure*}[t!]
\centering
\subfloat[]{\includegraphics[width = 0.213\textwidth]{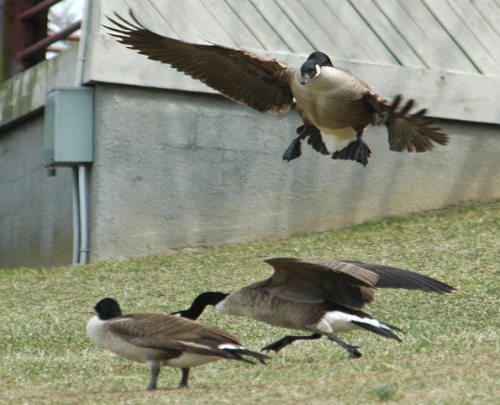}
\label{fig_ex_a}}
\hfil
\subfloat[]{\includegraphics[width = 0.213\textwidth]{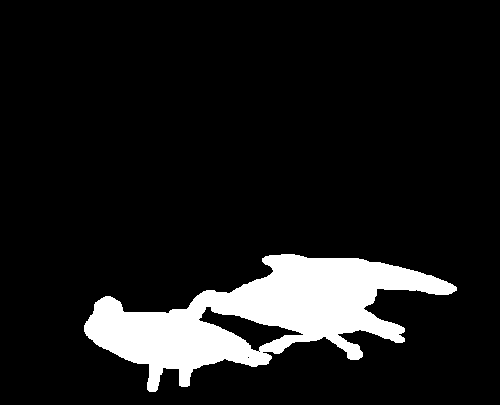}
\label{fig_ex_b}}
\hfil
\subfloat[]{\includegraphics[width = 0.213\textwidth]{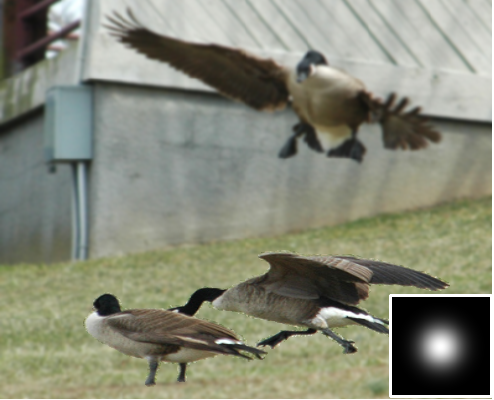}
\label{fig_ex_c}}
\hfil
\subfloat[]{\includegraphics[width = 0.213\textwidth]{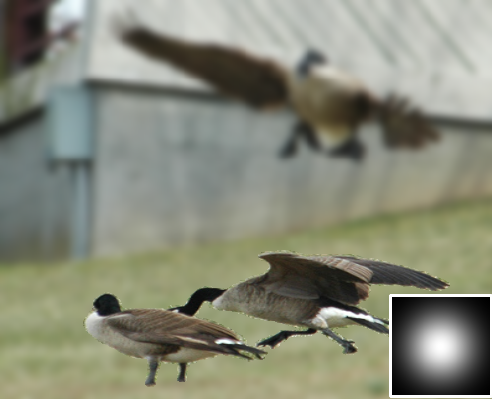}
\label{fig_ex_d}}

\subfloat[]{\includegraphics[width = 0.213\textwidth]{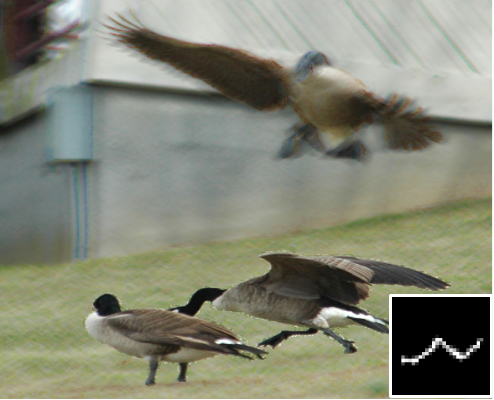}
\label{fig_ex_e}}
\hfil
\subfloat[]{\includegraphics[width = 0.213\textwidth]{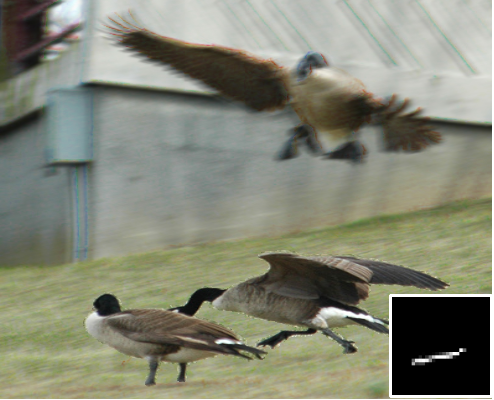}
\label{fig_ex_f}}
\hfil
\subfloat[]{\includegraphics[width = 0.213\textwidth]{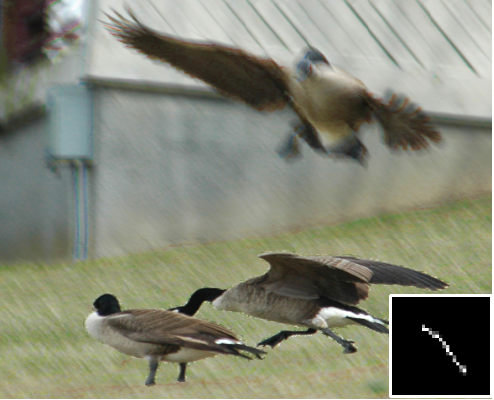}
\label{fig_ex_g}}
\hfil
\subfloat[]{\includegraphics[width = 0.213\textwidth]{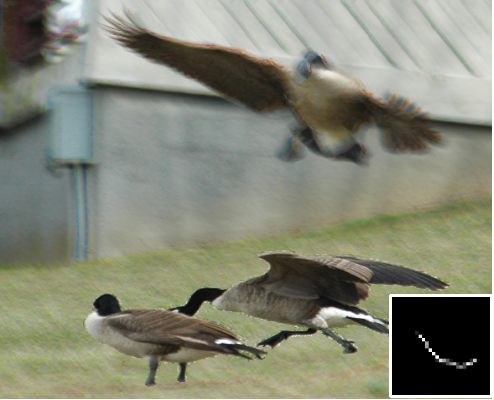}
\label{fig_ex_h}}
\caption{(a) Input image from the VOC dataset (b) Segmented Foreground (c-d) Blurring of input image with Gaussian blur, $\sigma=1.5,3$ pixels (e-h) Blurring of input image with randomly-generated non-linear motion blur. Blur kernels displayed at the right-bottom on each image. All images have been blurred with the halo-artifact removal described in section \ref{ssec:inpainting}. Blur differences are better appreciated when focusing in the roof on the top of the image.}
\label{fig_ex}
\end{figure*}

%% file: fig_halo.tex
\begin{figure*}[ht!]
\centering
\subfloat[]{\includegraphics[width = 0.15\textwidth]{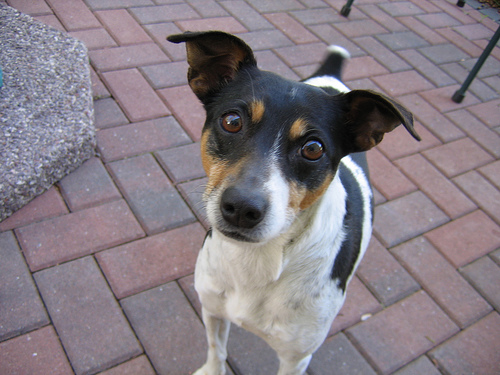}
\label{fig_blur1}}
\hfil
\subfloat[]{\includegraphics[width = 0.15\textwidth]{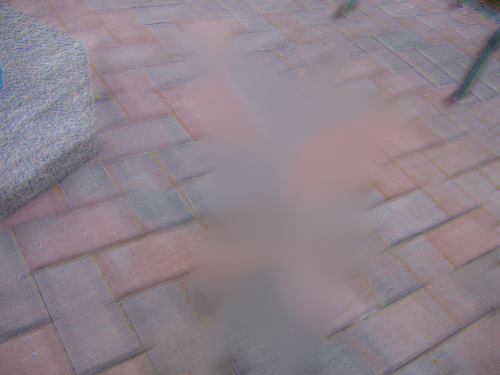}
\label{fig_blur2}}
\hfil
\subfloat[]{\includegraphics[width = 0.15\textwidth]{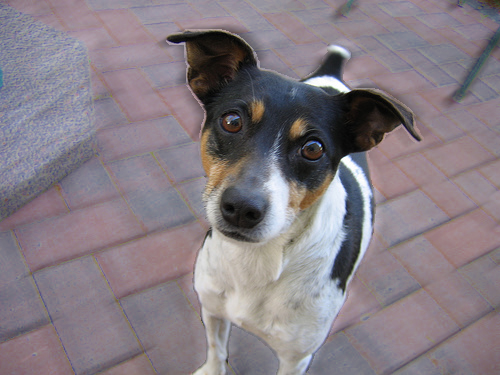}
\label{fig_blur3}}
\hfil
\subfloat[]{\includegraphics[width = 0.15\textwidth]{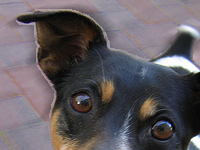}
\label{fig_blur4}}
\hfil
\subfloat[]{\includegraphics[width = 0.15\textwidth]{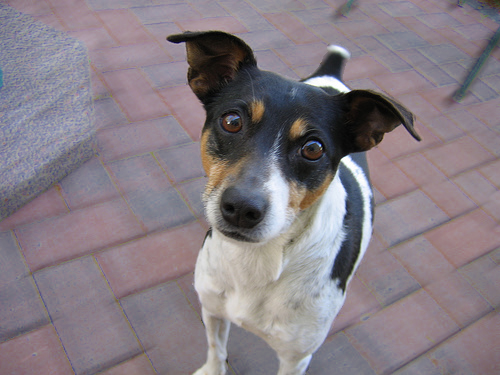}
\label{fig_blur5}}
\hfil
\subfloat[]{\includegraphics[width = 0.15\textwidth]{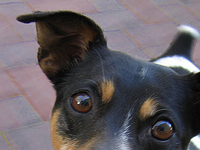}
\label{fig_blur6}}
\caption{(a) Original image from the VOC dataset (b) Inpainted foreground (c,d) Naively blurred background (e,f) Result of blurring after inpainting background.}
\label{fig_halo}
\end{figure*}

%% file: 3_model.tex
\section{Convolutional Neural Networks for Blur Segmentation from Synthetic Data}\label{sec:model}

\subsection{Architecture}\label{ssec:arch}

Section \ref{sec:synth} provides us with all the necessary tools to create a procedural -and thus, highly variable- training image flow. 
By posing blur localization as a dense, per-pixel classification task, we can make use of one of the well-established set of deep architectures devoted to semantic segmentation and feed it with such synthetically distorted image stream for training.
We consciously avoid the \emph{ad hoc} design of an architecture specifically suited for our target task, and instead rely on an off-the-shelf deep network, which allows us to isolate the contribution derived from the proposed training procedure.

We select the DeepLabv3\cite{chen_rethinking_2017} Network  as our reference CNN architecture, since it has recently shown state of the art performance on various dense labeling tasks.
This is partly due to the fact that it features an effective receptive field significantly larger than those of other standard architectures.
These are enabled by the use of \emph{atrous} convolutions (first introduced in this context in~\cite{chen_semantic_2015, yu_multiscale_2016}), which also contribute to the attainment of output maps with a large spatial resolution.
It has been previously shown that high semantic level features can greatly contribute to solve the task of blur localization~\cite{ma_deep_2018a}.
Hence, we can expect that the use of wide receptive fields, covering significant parts of the scene, will bring important benefits to our technique.
The DeepLabv3 architecture also benefits from the fusion of feature maps from multiple scales (which has proven crucial for solving the scale ambiguity problem~\cite{kim_defocus_2018}): the use of a new \emph{Atrous Spatial Pyramid Pooling} (ASPP) module helps capture context at various ranges.
In this work, we employ the DeepLabv3-ResNet101 variation of the architecture, which is constructed by a Deeplabv3 model with a ResNet-101 backbone.
We start from a model initially pre-trained on a subset of the COCO train2017 dataset~\cite{lin_microsoft_2014a} containing the $20$ classes also present in the Pascal VOC $2012$ segmentation dataset, and we substitute its final classifier with a $1\times1\times256\times2$ $2D$ convolution, followed by a $logSoftmax$ operation.

\subsection{Training procedure}\label{ssec:training}

Our training set for the purely self-supervised and weakly supervised approaches is generated by randomly applying synthetic defocus and motion blurring operations (generated as described in section \ref{sec:synth}) to image batches (with a batch size of 18) from the training set of the SegVOC12 dataset~\cite{everingham_pascal_2015}.
Such degradations are selectively applied according to the binary mask created on the fly as mentioned in sections \ref{sssec:semantic_object_masks} and \ref{sssec:object_proposals} for the ground truth semantic object masks and object proposal-based mask, respectively, and the blur mask itself is used as the ground truth label.

Additionally, the usual standard random data transformations (affine transforms, flips, color jitter, cropping) are applied to the input and target image pairs before resizing them from their original size to $224\times224$ pixels.
A randomly applied JPEG compression-based augmentation is also performed at different stages of the input image preprocessing workflow, with the aim of gaining invariance to low level, dataset-specific regularities.

We employ a negative log-likelihood loss that we minimize using the Adam optimizer, and let the model train until the validation loss stagnates for 20 epochs.
A reduced number of hyperparameter tuning configurations was tried for each of the experimental setting, and the setup yielding the lowest validation loss was kept for evaluation.
In most of the configurations, this corresponded to a learning rate value of $10^{-5}$ and a weight decay of $5\cdot10^{-4}$.

%% file: 4_results.tex
\section{Experimental Results}\label{sec:results}

We evaluate the trained model on the largest publicly available\footnote{Zhang \etal's dataset~\cite{zhang_learning_2018} and Zhao \etal's supplementary train set~\cite{zhao_defocus_2019} are not open.} dataset of images annotated in terms of defocus and object-motion blur localization, i.e. that of Shi \etal~\cite{shi_discriminative_2014}\footnote{Code and models are available on \href{https://github.com/aitorshuffle/synthblur}{https://github.com/aitorshuffle/synthblur}.}.
The dataset consists of $1000$ partially blurred natural images with human-made binary (blur/no-blur) pixel-wise annotations, of which $704$ correspond to defocus blur and $296$ to motion blur-affected images.
Following \cite{ma_deep_2018a}, we partition the dataset in an odd and an even subset, both of them containing an approximately equal amount of images affected by both types of blurs), and keep the latter held out for testing purposes.
Unless otherwise stated, all the results shown in this section were thus evaluated on the $500$ images from Shi \etal's even subset.
Meanwhile, a $20\%$ of the odd subset ($100$ images) is used for validation, and the remaining $400$ images are used for training in those experimental setups that require a supervised training component (i.e. our semi-supervised approach and the fully supervised model from Table~\ref{tab:results_shi2014} and Fig~\ref{fig_perf_series_lineplot}).

At inference, we run a simple test-time augmentation (TTA) process, consisting of averaging the predictions yielded by the original input image, together with its horizontally flipped version, and upscale the resulting $224\times224$ pixels-sized predictions to the original, varying image sizes before performing the evaluation. 
Even though the raw performance values could, to some extent, benefit from performing a re-scaling of the resulting blur map in the $[0,1]$ range, we purposefully avoid such kind of post-processing, in order to enable the prediction of completely sharp or blurred scenes.
Consequently, we can consider the values of the predicted mask pixels as a measure of absolute blurriness, as opposed to a measure of blur level of each region with respect to the sharpest area of the image.

\subsection{Self-supervised setup}\label{ssec:selfsup_setup}

We first consider the purely self-supervised instantiation of our framework, in which we directly test the model trained on synthetically blurred images over the $500$ even samples of the test dataset.
We compare our method against most of the best performing hand-crafted feature-based approaches which do not require any dedicated dataset for training: 
Liu \etal~\cite{liu_image_2008}, Chakrabarti \etal~\cite{chakrabarti_analyzing_2010}, Su \etal~\cite{su_blurred_2011}, Shi \etal~\cite{shi_discriminative_2014}, LBP~\cite{yi_lbp-based_2016} and HiFST~\cite{golestaneh_spatially-varying_2017}.
In addition, we include the performance reported for the same even subset by one of the most recent deep CNN-based defocus and motion blur detection methods in the literature, \ie, the \emph{Deep Blur Mapping} approach by Ma \etal~\cite{ma_deep_2018a}.

\input{table_results_shi2014}

Table~\ref{tab:results_shi2014} shows the obtained results in terms of Area Under the Receiver Operating Characteristic (ROC) Curve (AUC) and Average Precision (AP), as computed over the different values of recall given by the Precision-Recall (P-R) curve.
Both the AUC and AP values were computed individually for each output map and then averaged over the whole even test subset.
The overall performance value (\emph{All}) is also shown disaggregated for the \emph{Defocus} and \emph{Motion} subsets of the database.
Note that, as is the case with most of the other methods, in absolute terms our approach performs better over the defocused images than over the motion-blurred ones.
This will hold true for the results of all the three variants of our framework.

From Table~\ref{tab:results_shi2014} we can see that our self-supervised approach performs significantly better than all the other non-deep methods for every metric and blur-type subset. 
Furthermore, the proposed self-supervised learning method, when used to train the off-the-shelf DeepLabv3\_resnet101 network and without ever observing a single image with real blur, yields better overall AUC and AP values than Ma \etal's CNN architecture~\cite{ma_deep_2018a}, whose design was tuned \emph{ad hoc} for this task and trained end-to-end in a fully supervised setup over the $500$ odd samples of the dataset\footnote{The performance evaluation protocol in \cite{ma_deep_2018a} is slightly different, as they compute a single AP value for a one-dimensional vector containing the predictions of all the pixels of every image in the even subset. Although we believe that the AUC/AP values should be computed on a per-image basis, under their protocol our self-supervised overall AP is $0.952$, vs. their reported value of $0.880$}.
In particular, our method performs close to Ma \etal's on the defocus blur subset, but significantly better for the motion blur samples.
The last row of Table~\ref{tab:results_shi2014} (\ie \emph{Fully supervised}) shows the results obtained by fine-tuning the DeepLabv3\_resnet101 network over Shi's odd subset.
The mixed results of this fully supervised model when compared to Ma \etal suggest that the performance gain of our self-supervised method is largely due to the benefits of our training scheme, rather than being just a product of the use of a better architecture. 

Other relevant fully supervised CNN based results were left out of this comparison for various reasons: 
Zhao \etal~\cite{zhao_defocus_2018} target exclusively images degraded with defocus blur, and their training procedure involves 
\begin{enumerate*}[label=(\roman*)]
    \item pre-training the model employing very simple artificial defocus blurring operations over half of the image on samples from additional datasets and 
    \item further fine-tuning it over $604$ out of the $704$ images of Shi \etal's defocus blur partition and testing it over the remaining $100$.
\end{enumerate*}
Finally, Zhang \etal~\cite{zhang_learning_2018} train their \emph{ad hoc} designed \emph{ABC-FuseNet} architecture end-to-end on their unreleased \emph{SmartBlur} blur segmentation dataset of $10.000$ images before evaluating on Shi \etal. 
Even with such amount of training samples, their reported AP is $0.869$ for the whole dataset, sensibly below our results on the even partition.

\input{fig_res_qual}
Fig.\ref{fig_res_qual} contains visual results for a small random subset of images affected by both types of blur (defocus blur in the top seven rows, motion blur in the bottom seven), as predicted for most of the considered methods.
We can observe that, even without the utilization of any single ground truth blur segmentation annotation from the target dataset for direct supervision, our self-supervised approach obtains accurate masks, comparable in visual quality to those produced by fully-supervised deep CNN-based methods, such as~\cite{ma_deep_2018a}.

\input{table_results_shi2014_blur_type_ablation}
\paragraph{Blur type ablation} Finally, Table~\ref{tab:results_shi2014_blur_type_ablation} shows the results obtained over the same sets when only defocus blur or only motion blur synthetic degradations where applied during training in our self-supervised setup. 
The results suggest that devoting the full capacity of the network to learning to detect only one specific type of degradation does help in the case of defocus blur training, but not so when the training is constrained to observing samples affected solely by object motion.
For the latter, the increase of variability introduced by feeding the net with both kinds of degradations seems beneficial, probably due to a regularization effect.
Cross-blur type evaluation is asymmetric: while training on motion-only blur achieves decent results on the defocus-only test subset, a model trained uniquely on defocus blur synthetic degradations suffers a significant performance degradation if applied to object motion-affected images.

In addition, this experiment reveals one of the advantages inherent to our blur detection framework: by selectively tunning the ratio of images being synthetically blurred with defocus or motion kernels during training, we can prioritize the performance over either kind of blurs, or operate anywhere in between.

\subsection{Weakly supervised setup}\label{ssec:weaksup_setup}

The \emph{Ours weakly supervised}-labeled row from Table \ref{tab:results_shi2014} shows the results achieved with our weakly supervised approach, based on the use of manually annotated semantic object segments as blur masks instead of the unsupervised proposals yielded by the MCG algorithm.
While one could intuitively think of this as an upper bound for our self-supervised method's performance, we observe that, in fact, both methods perform almost on pair, with the self-supervised approach yielding slightly better results for the images affected by defocus blur and with no clear strongest method in terms of motion-blur and overall results, depending on the metric of interest.
This experiment shows that applying blur degradations to object proposals computed with an object proposal method does not negatively impact our results, and that similar outcomes are obtained when compared to a method trained with ground truth segmentation masks.

\subsection{Semi-supervised setup}\label{ssec:semisup_setup}

\input{fig_perf_series_lineplot}

We now introduce a modification in the training process in order to test the usefulness of our proposed synthetically blurred image-based training when a limited number of blur segmentation annotations are available for the target dataset.
This corresponds to our semi-supervised experimental setting, in which a joint training is performed: 
mini-batches are now composed of equal amount of image and ground truth blur mask pairs produced (i) in a MCG-based synthetic blurring operation, and (ii) sampled from the train subset of the target dataset~\cite{shi_discriminative_2014}.

In order to assess the usefulness of this semi-supervised setting, we consider the $400$ images from Shi \etal's training set that were previously separated and employ a varying fraction of them for joint training with the synthetically blurred images. 
Specifically, we conduct the experiment with a $2\%$, $4\%$, $5\%$, $10\%$, $20\%$, $40\%$ and $80\%$ of the odd part being used as training aid, which, in absolute terms, correspond to $10$, $20$, $25$, $50$, $100$, $200$, and $400$ images, respectively.
The evaluation protocol is not affected, and the trained model is then tested on the $500$ even pairs of the dataset.

We compare our semi-supervised results with those presented in the preceding sections (i.e. \emph{Ours self-supervised} and \emph{Ours weakly supervised} approaches from Table~\ref{tab:results_shi2014}), whose performance metrics do not vary with the number of real images being used for joint training.
Finally, for comparison we extend the fully supervised approach presented in section~\ref{ssec:selfsup_setup} by fine-tuning the same deeplabv3\_resnet101 model over the same fractions of the odd train partition.

Fig.~\ref{fig_perf_series_lineplot} shows the AUC and AP values obtained for each of the aforementioned methods, both as overall metrics and with disaggregated values for defocus and motion blur affected images.
We observe that: 
\begin{itemize}
	\item For all the considered amounts of annotated images, our joint, semi-supervised approach outperforms the fully supervised one by a significant margin for both blur types, especially so as we operate with few real samples.
	This means that the use of synthetically generated blurred image-ground truth pairs has proven useful as additional source of training data for improving the performance of fully supervised blur detection approaches trained end-to-end.
	\item The concrete AUC and AP values obtained by these variations for $400$ real blur images were also added, for comparison, to Table~\ref{tab:results_shi2014}.
	As shown there, the semi-supervised training scheme achieves the best overall results and, for every subset and metric, there is always one of the three instantiations of our framework outperforming every other deep fully supervised alternative.
	In particular, our best method (semi-supervised setting) reaches an overall AUC of $0.941$ and an AP of $0.934$, $0.019$ and $0.022$ points better than Ma \etal's, respectively.
	\item Even the self-supervised or the technically simpler but more annotation-dependent weakly supervised variants of our framework can clearly outperform both their semi-supervised counterpart and, most notably, the fully supervised training in the lower part of the range.
	\item This gap is closed by the fully supervised training scheme only for the defocus blur subset, as we keep adding images with real blur, but not so in the case of the motion blur subset.
\end{itemize}

\subsection{Cross-dataset generalization}\label{ssec:cross_dataset}

One of the most frequent limitations of current end-to-end trained CNN-based solutions to many computer vision tasks is the inability of models trained on a certain dataset to generalize to other datasets with some underlying distribution shift, either revealed as some readily apparent visual difference (e.g. illumination, object appearance) or due to some hard-to-perceive low level statistical regularities within the dataset.
Domain adaptation solutions~\cite{tsai_learning_2018} aim at mitigating the harmful effect of such domain shifts, but they are frequently based on domain-adversarial training strategies that can be cumbersome to implement, and mostly ineffective in providing a significant performance gain when both domains are easily told apart.

The following experiment seeks to evaluate this cross-dataset generalization ability for our self-supervised approach as compared to that of other methods.
To that end, we introduce the dataset of defocus blurred images provided by Zhao \etal in \cite{zhao_defocus_2018}, which will serve for testing purposes. 
\input{table_results_zhao2018}
Table \ref{tab:results_zhao2018} shows the results obtained by Zhao \etal themselves, along with Ma \etal's and the aforementioned fully supervised DeepLabv3 model.
The reported values are all result of the direct application of the models learned over Shi \etal's dataset.
These are further compared with our weakly-supervised and self-supervised training methods, showing that the latter exhibits a significantly better generalization ability ($0.950$ vs. $0.923$ of AUC for Ma \etal's approach).

As a general conclusion, we show that synthetically blurring parts of images with a certain semantic coherence is, on its own, a useful technique to perform self-supervised or weakly supervised blur localization, and that its use as aid when performing supervised end-to-end training (even in extreme few-shot cases) can help boost detection accuracy.
Those domains where annotated data is scarce can particularly benefit from such approaches when applied in conjunction with other downstream computer vision tasks.
Histological imaging (to focus a potential disease classification model on sharp parts of a digitized slide), multispectral imaging (where a significant, hardly avoidable blur effect is often found due to the chromatic aberrations derived from the large bandwidth of the captured spectra, and it is difficult to tell it apart from motion or defocus blur effects) or document scanning are some examples of such situations where a self-supervised approach could have a large impact.

%% file: table_results_shi2014.tex
\begin{table*}[!htpb]  
	\centering
	\sisetup{detect-weight=true,detect-inline-weight=math}
	\begin{tabular}{l ccc ccc}
		\toprule
		& \multicolumn{3}{c}{AUC} & \multicolumn{3}{c}{AP} \\
		\cmidrule(lr){2-4} \cmidrule(lr){5-7} 
		Method                                              & Defocus & Motion & All & Defocus & Motion & All\\
		\midrule
		Liu \etal\cite{liu_image_2008}                      & 0.722 & 0.714 & 0.720 & 0.792 & 0.683 & 0.760 \\ 
		Chakrabarti~\cite{chakrabarti_analyzing_2010}		& 0.745 & 0.640 & 0.714 & 0.837 & 0.675 & 0.789 \\ 
		Su \etal\cite{su_blurred_2011}                      & 0.807 & 0.750 & 0.790 & 0.859 & 0.707 & 0.814 \\ 
		Shi \etal\cite{shi_discriminative_2014}             & 0.836 & 0.735 & 0.806 & 0.876 & 0.699 & 0.823 \\ 
		LBP\cite{yi_lbp-based_2016}                   		& 0.855 & 0.678 & 0.802 & 0.876 & 0.683 & 0.819 \\ 
		HiFST\cite{golestaneh_spatially-varying_2017} 		& 0.901 & 0.804 & 0.873 & 0.928 & 0.744 & 0.874 \\ 
		\midrule
		Ma \etal\cite{ma_deep_2018a}             		    & \colsecond{0.947} & 0.861 & 0.922 & \colsecond{0.966} & 0.784 & 0.912   \\ 
		\midrule
		Ours self-supervised                                & \colthird{0.945} & \colfirst{\underline{0.905}} & \colsecond{0.933} & 0.960 & \colthird{0.838} & \colthird{0.924} \\ 
        Ours weakly supervised (segmentation masks)         & 0.941 & \colthird{0.897} & \colthird{0.928} & 0.959 & \colfirst{\underline{0.849}} & \colsecond{0.926} \\ 
		Ours semi-supervised (joint with 400 odd img.) & \colfirst{\underline{0.956}} & \colsecond{0.904} & \colfirst{\underline{0.941}} & \colfirst{\underline{0.974}} & \colsecond{0.840} & \colfirst{\underline{0.934}} \\  
		\midrule
		Fully supervised (finetuned to 400 odd img.) & 0.943 & 0.875 & 0.923 & \colthird{0.965} & 0.819 & 0.922 \\ 
 		\bottomrule
	\end{tabular}
	\caption{Quantitative evaluation over Shi \etal's dataset's~\cite{shi_discriminative_2014} even partition. \colfirst{\underline{Best}}, \colsecond{2nd best} and \colthird{3rd best} results are highlighted for each metric and blur type.}
	\label{tab:results_shi2014}
\end{table*}

%% file: fig_res_qual.tex
\captionsetup[subfigure]{labelformat=empty}
\begin{figure*}[!htpb]
    \centering
	\subfloat[Input]{\includegraphics[width=0.0995\textwidth]{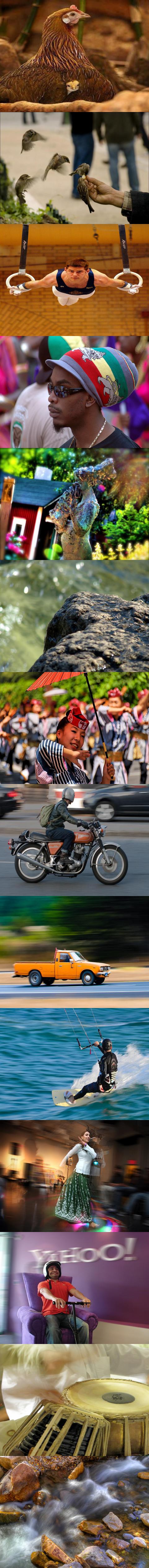}\label{fig_res_qual_a}}
	\hfil
	\subfloat[Ground truth]{\includegraphics[width=0.0995\textwidth]{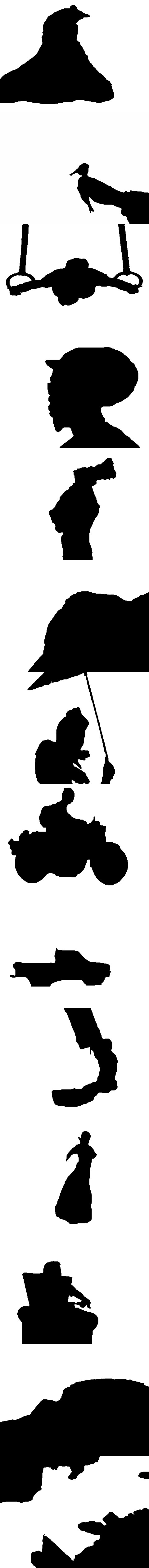}\label{fig_res_qual_b}}
	\hfil
	\subfloat[Liu\cite{liu_image_2008}]{\includegraphics[width=0.0995\textwidth]{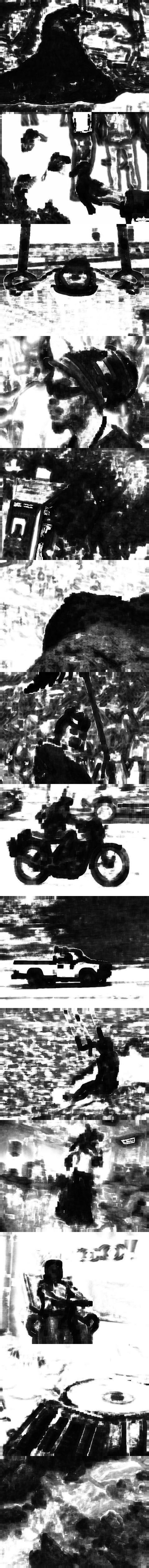}\label{fig_res_qual_c}}	
	\hfil
	\subfloat[Chakraba.\cite{chakrabarti_analyzing_2010}]{\includegraphics[width=0.0995\textwidth]{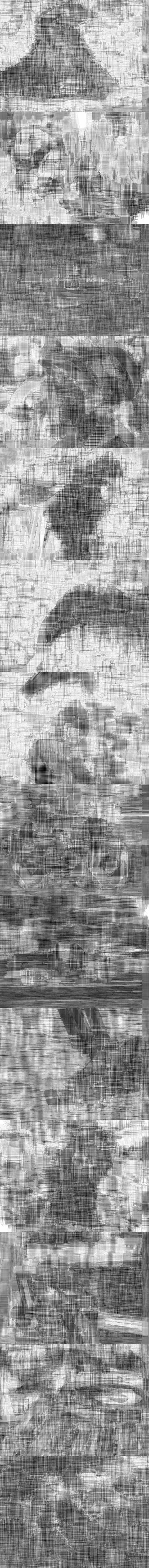}\label{fig_res_qual_d}}
	\hfil
	\subfloat[Su\cite{su_blurred_2011}]{\includegraphics[width=0.0995\textwidth]{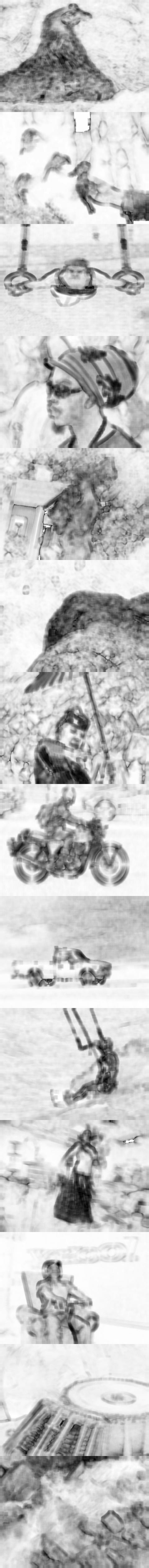}\label{fig_res_qual_e}}	
	\hfil
	\subfloat[Shi\cite{shi_discriminative_2014}]{\includegraphics[width=0.0995\textwidth]{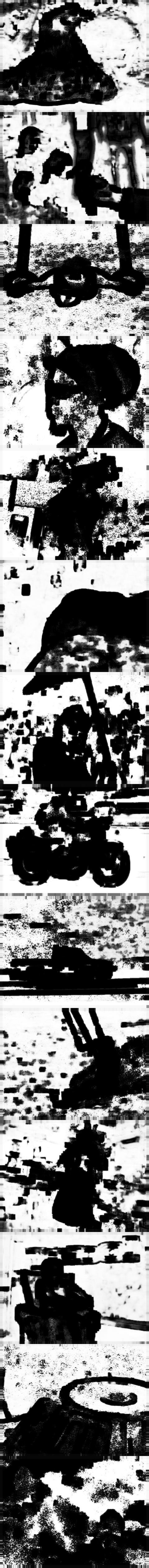}\label{fig_res_qual_f}}	
	\hfil
	\subfloat[LBP\cite{yi_lbp-based_2016}]{\includegraphics[width=0.0995\textwidth]{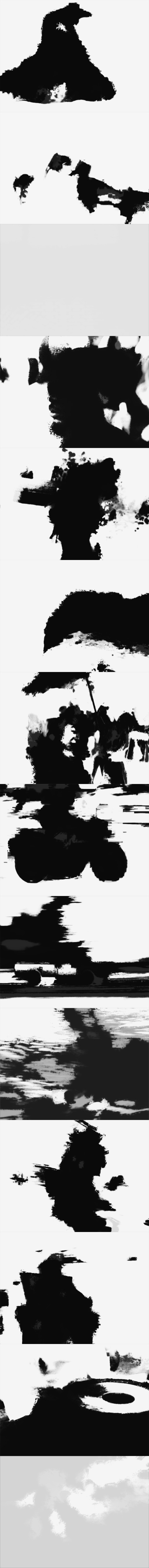}\label{fig_res_qual_g}}	
	\hfil
	\subfloat[HiFST\cite{golestaneh_spatially-varying_2017}]{\includegraphics[width=0.0995\textwidth]{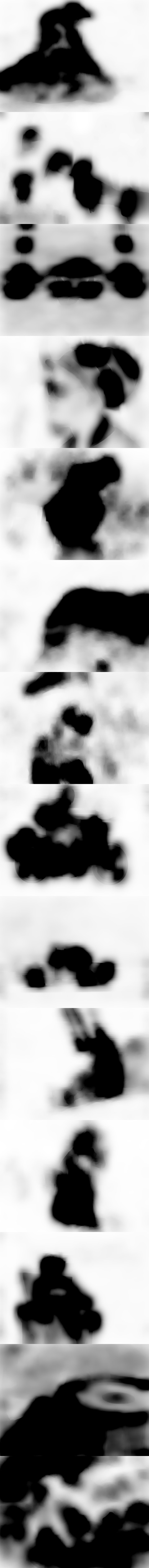}\label{fig_res_qual_h}}	
	\hfil
	\subfloat[Ma\cite{ma_deep_2018a}]{\includegraphics[width=0.0995\textwidth]{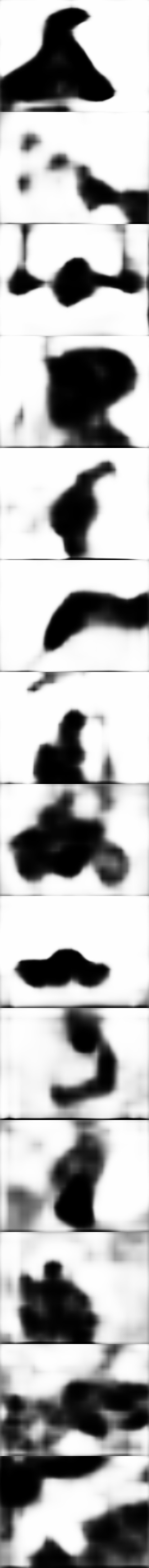}\label{fig_res_qual_i}}	
	\hfil
	\subfloat[Ours self-sup.]{\includegraphics[width=0.0995\textwidth]{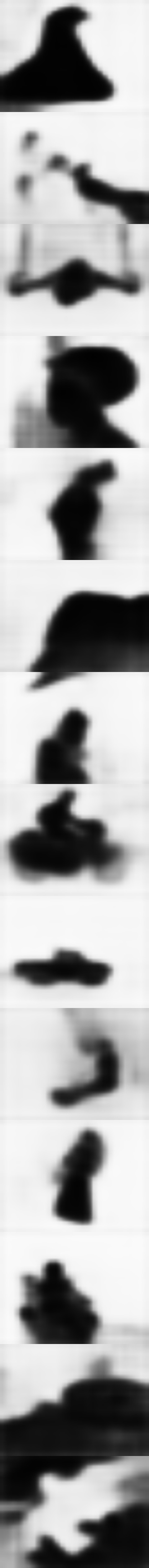}\label{fig_res_qual_j}}	
	\caption{Qualitative results for a sample of images from Shi's dataset\cite{shi_discriminative_2014} affected by defocus (top $7$) and motion (bottom $7$) blur processed by the different evaluated algorithms.}
	\label{fig_res_qual}
\end{figure*}
\captionsetup[subfigure]{labelformat=parens} 

%% file: table_results_shi2014_blur_type_ablation.tex
\begin{table}[!htpb]  
	\centering
	\sisetup{detect-weight=true,detect-inline-weight=math}
	\begin{tabular}{l ccc ccc}
		\toprule
		& \multicolumn{3}{c}{AUC} & \multicolumn{3}{c}{AP} \\
		\cmidrule(lr){2-4} \cmidrule(lr){5-7} 
		 & DF & MT & All & DF & MT & All\\
		\midrule
		DF  & \textbf{0.949}& 0.826         & 0.913         & \textbf{0.967}& 0.773         & 0.910         \\
		MT  & 0.934         & 0.894         & 0.922         & 0.953         & 0.831         & 0.917         \\ 
		All & 0.945         & \textbf{0.905}& \textbf{0.933}& 0.960         & \textbf{0.838}& \textbf{0.924}\\ 
 		\bottomrule
	\end{tabular}
	\caption{Blur type based ablation test over Shi \etal's dataset's~\cite{shi_discriminative_2014} even partition, in our self-supervised setup. Rows represent the synthetic blur type being applied on training (DF=Defocus, MT=Motion, All=Defocus and Motion). Columns represent the test (sub)set. Bold is best.}
	\label{tab:results_shi2014_blur_type_ablation}
\end{table}

%% file: fig_perf_series_lineplot.tex

\begin{figure*}[t]
	\centering
	\subfloat[]{\includegraphics[width = 1.0\textwidth]{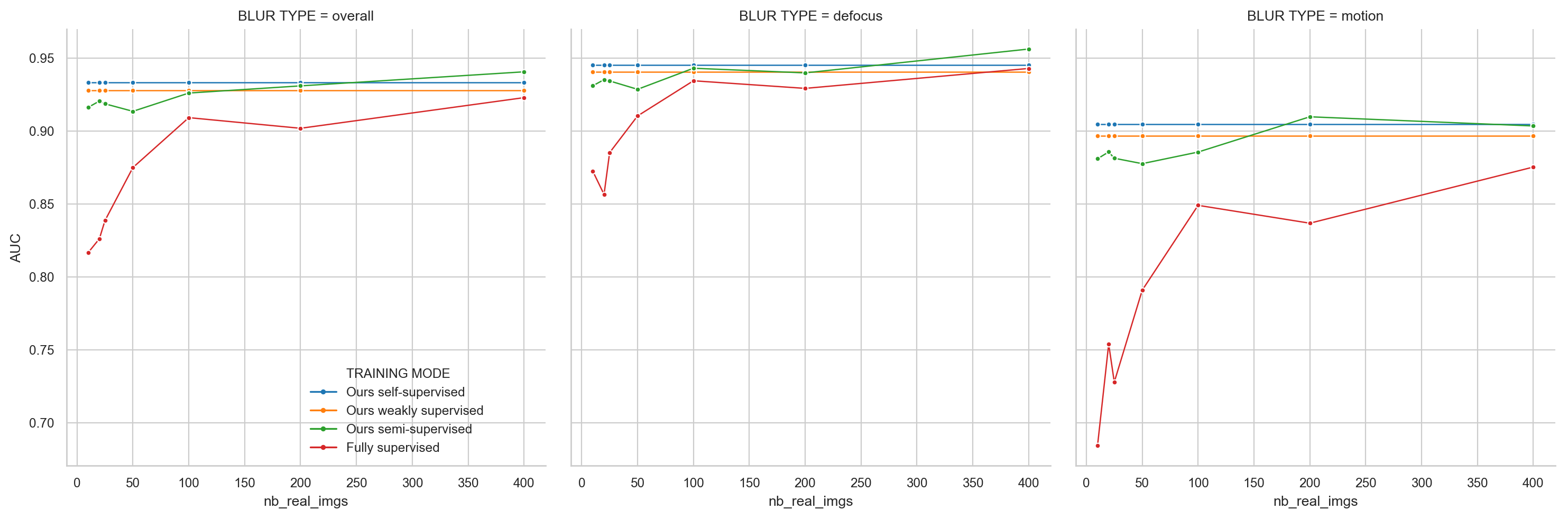}
		\label{fig:shi2014_auc}}
	\hfil
	\subfloat[]{\includegraphics[width = 1.0\textwidth]{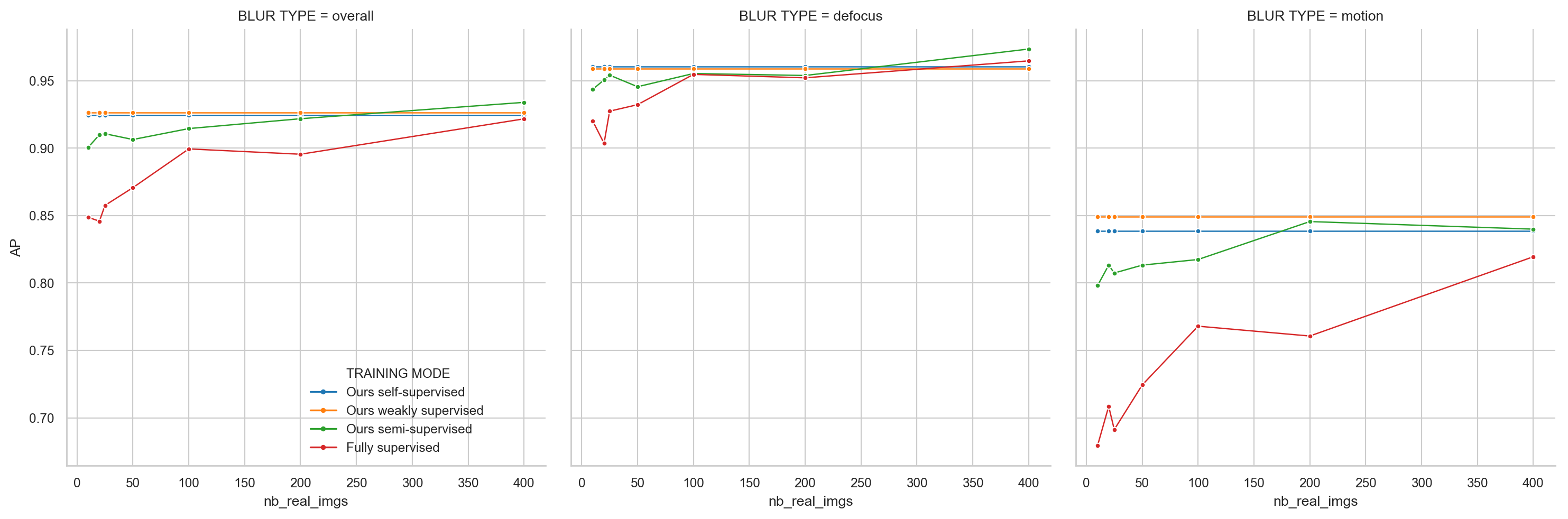}
		\label{fig:shi2014_ap}}
	\caption{(a) AUC and (b) AP as a function of the number of images with real blur from the Shi \etal's dataset~\cite{shi_discriminative_2014} used in the training process, in the following setups:
	(i) Joint training on images with synthetic and real blur (\emph{Ours semi-supervised.})
	(ii) Direct fully-supervised fine-tuning on images from Shi \etal (\emph{Fully supervised}).
	The following setups are shown for comparative purposes, but do not use any image from Shi \etal:
	(iii) MCG object proposals-based self-supervised training (\emph{Ours self-supervised}), 
	(iv) SegVOC12 semantic segmentation masks-based weakly supervised training (\emph{Ours weakly supervised}).
	The set of images with real blur ar part of Shi \etal's odd subset, containing both defocus and motion blur.
	All the experiments were done using the DeepLabv3 architecture~\cite{chen_rethinking_2017} with a Resnet101 backbone.}
	\label{fig_perf_series_lineplot}
\end{figure*}

%% file: table_results_zhao2018.tex
\begin{table}[!htpb]
	\centering
	\sisetup{detect-weight=true,detect-inline-weight=math}
	\begin{tabular}{l c c}
		\toprule
		Method & AUC & AP\\
		\midrule
		Zhao \etal~\cite{zhao_defocus_2018} & 0.913 & 0.946 \\ 
		Ma \etal~\cite{ma_deep_2018a}       & 0.923 & 0.956 \\ 
		\midrule
 		Ours self-supervised                    & \textbf{0.950} & \textbf{0.976} \\
 		Ours weakly supervised                  & 0.915 & 0.953 \\	
 		\midrule
 		Fully supervised                        & 0.904 & 0.952 \\ 
		\bottomrule
	\end{tabular}
	\caption{Direct testing of models from Table~\ref{tab:results_shi2014} on Zhao \etal's defocus blur dataset~\cite{zhao_defocus_2018}, together with Zhao \etal's ~\cite{zhao_defocus_2018} own results. None of the models in this table have seen Zhao \etal's dataset during training. Bold is best.}
	\label{tab:results_zhao2018}
\end{table}

%% file: 5_conclusions.tex
\section{Conclusions}\label{sec:conclusions}

This paper presents a framework for deep defocus and object motion blur segmentation built upon the procedural application of both types of synthetic blurring distortions over regions of images.
The self-supervised and weakly supervised versions of the framework exploit different ways of obtaining the candidate blur masks for automatic ground truth generation, and can be applied without any blur localization annotation.
In the semi-supervised case, this source of data augmentation can leverage the availability of a few labeled images to further improve the obtained segmentation accuracies.
Extensive quantitative and qualitative experiments show that a segmentation CNN trained on this kind of synthetic data under any of the three mentioned framework configurations is able to accurately localize the blurred regions of a hold-out set, showing performances well above other recent CNN-based approaches.

%% file: 6_acknowledgements.tex
\section{Acknowledgements}\label{sec:acknowledgements}
This research was partially funded by the Basque Government's Industry Department under the ELKARTEK program's project ONKOIKER under agreement KK-2018/00090.
We thank the Spanish project TIN2016-79717-R and mention Generalitat de Catalunya CERCA Program. 
\newpage